# Can mobile usage predict illiteracy in a developing country?

Pål Sundsøy

Telenor Group Research, Big Data Analytics
Snarøyveien 30,1331 Fornebu, Norway

**Abstract.** The present study provides the first evidence that illiteracy can be reliably predicted from standard mobile phone logs. By deriving a broad set of mobile phone indicators reflecting users' financial, social and mobility patterns we show how supervised machine learning can be used to predict individual illiteracy in an Asian developing country, externally validated against a large-scale survey. On average the model performs 10 times better than random guessing with a 70% accuracy. Further we show how individual illiteracy can be aggregated and mapped geographically at cell tower resolution. Geographical mapping of illiteracy is crucial to know where the illiterate people are, and where to put in resources. In underdeveloped countries such mappings are often based on out-dated household surveys with low spatial and temporal resolution. One in five people worldwide struggle with illiteracy, and it is estimated that illiteracy costs the global economy more than $1 trillion dollars each year [*1*]. These results potentially enable cost-effective, questionnaire-free investigation of illiteracy-related questions on an unprecedented scale.

## 1 Introduction

Functional illiteracy means a person may be able to write simple words, but cannot apply these skills to tasks such as filling out a job application, reading a medicine label or balancing a chequebook [*2*]. Illiterates are often trapped in a cycle of poverty with limited opportunities for employment or income generation [*3*]. They also have higher chances of poor health, turning to crime and dependence on social welfare or charity when available [*4*]. Mapping of literacy statistics is currently based on tedious household surveys with a low spatial and temporal frequency [*5*]. The increasing availability and reliability of new data sources, and the growing demand of comprehensive, up-to-date international literacy data are therefore of high priority [*6*]. One of the most promising rich Big Data sources are mobile phone logs [*7*], which have the potential to deliver near real-time information of human behaviour on individual and societal scale [*8*]. Several research studies have used large-scale mobile phone metadata, in the form of call detail records (CDR) and airtime purchases (top-up) to quantify various socio-economic dimensions. Eagle et al. [*9*] quantified the correlation between network diversity and a population's economic well-being. The findings revealed that the diversity of individuals' relationships is strongly correlated with the economic development of communities. The assumption that more diverse ties correlate with better access to social and economic opportunities was untested at the population level. It concludes that frequently making and receiving calls with contacts outside of one's immediate community is correlated with higher socio-economic class. CDRs have also shown to provide proxy indicators for assessing regional poverty levels, as shown in studies done in several countries [*10,11,12*]. It has also been hypothesized that airtime purchases is correlated with socio-economic status [*13*], and this has later been extended and verified also using external reliable data [*14*] . Monitoring airtime transactions for trends and sudden changes can also be useful for detecting early impact of economic crisis [*15*], as well as for measuring impact of programmes designed to improve livelihoods and food security [*16*].

The rest of this paper is organized as follows: In section 2 we describe the methodological approach, including the features and modelling approach. In section 3 we describe the results. In chapter 4 we discuss the limitations from a holistic perspective, while we finally draw our conclusions in section 5.

## 2 Approach

We collect educational status, including illiteracy, from 76 000 individuals in a low HDI Asian country based on two large-scale surveys run by a professional agency on behalf of a large mobile operator. The sample includes 6.8% illiterates, 40% primary degree, 26% SSC, 17.6% HSC, 5.6% bachelor, 3.5% master and 0.13% other degrees (incl. Ph.D.). The survey is representative for all mobile users in the country, where more than 85% have a mobile phone. In general, half of the world's population now have a mobile phone [*17*]. The individual records are de-identified and coupled with 6-month raw mobile phone metadata, including CDR and airtime purchases. Content information, such as SMS content, is never accessed by the operator or any researcher. The educational status is re-labelled to a binary classifier, namely illiterates and non-illiterates. This section describes the features and the machine learning algorithm used for our prediction.

### 2.1 Features

A structured dataset consisting of 160 mobile phone features are built, and categorized into three dimensions: (1) financial (2) mobility and (3) social features, as shown in Table 1. The features are custom made and include various parameters of the corresponding distributions such as weekly or monthly median, mean and variance.

**Table 1.** Sample of features from mobile phone metadata used in model

| Dimension | Features |
|---|---|
| Financial 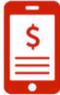 | **Airtime purchases:** Recharge amount per transaction, Spending speed, fraction of lowest/highest recharge amount, coefficient of variation recharge amount etc |
| | **Revenue:** Charge of outgoing/incoming SMS, MMS, voice, video, value added sevices, roaming, internet etc. |
| | **Handset:** Manufacturer,brand, camera enabled, smart/feature/basic phone etc |
| Mobility 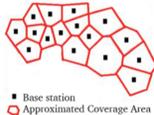 | Home district/tower, radius of gyration, entropy of places, number of places visited etc. |
| Social 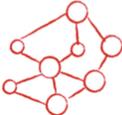 | **Social Network:** Interaction per contact, degree, entropy of contacts etc. |
| | **General phone usage:** Out/In voice duration, SMS count, Internet volume/count, MMS count, video count/duration, value added services duration/count etc. |

## 2.2 Model algorithm

Several model algorithms were tested such as neural networks, random forest and support vector machines. Based on performance, a gradient boosted machines model (GBM) ended up as the final model [*18*]. Here the base classifiers are built sequentially, and the algorithm combines the new classifier with ones from previous iterations in an attempt to reduce the overall error rate. The main motivation is to combine several weak models to produce a powerful ensemble. To compensate class imbalance, the minority class in the *training set*, that includes the illiterates being 6.8%, is up-sampled. The minority class is then randomly sampled, with replacement, to be the same size as the majority class. A 10-fold cross-validation is used as re-sampling technique. Feature importance scores are calculated via a backwards elimination feature selection routine that looks at reductions in the generalized cross-validation (GCV) estimate of error. It tracks the changes in GCV, for each predictor and accumulates the reduction in the statistic when each predictor's feature is added to the model. In our set-up, each model is trained and tested using a 75/25 split. All results are reported for the test-set.

## 3 Results

### 3.1 Individual illiteracy

Figure 1 shows the final features and their contribution in predicting illiteracy. Concretely, 19 of our features were related to illiteracy and were all included in the final GBM classifier. The model predicted whether phone users were illiterate with an accuracy of 70.1% (95% CI: 69.6-70.8). The deviation of accuracy from the training set was only 3.8%, which disregard model overfitting. The true positive rate (sensitivity/recall) was 71.6% and true negative rate (specificity) 70%. Given the original baseline of 6.8% in the test set, we predict on average 10 times better than random.

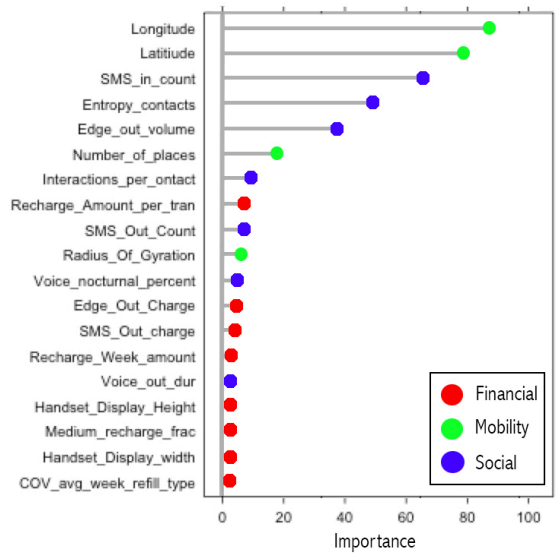

**Fig. 1.** Top features in the GBM Model colored by their respective feature family

An investigation of the most important predictors, as seen in Fig 1, reveals some interesting associations. We especially notice that most frequently used longitude and latitude stand out as good predictors: where the people spend most of their time is a good signal of their education level. One explanation can be that this signal indicates that the model catch regions of low economic development status, e.g. slum areas where illiteracy is high. Another important feature is the number of incoming SMS, which outperforms outgoing SMS: one hypothesis is that people do not send SMS' to persons that they know are illiterate. Moreover, we see that entropy of contacts is important – illiterates tend to concentrate their communication on few people. This is also in line with Eagle's work on geographical level [9], which shows that economic well-being is correlated with social diversity. Further we see that illiterates have limited use of internet (predictor 5), and their mobility pattern is limited to a few base stations (predictor 6).

The corresponding density distributions of the top 6 predictors are shown in Fig 2. We notice sharp peaks in the location distributions (Fig 2 A and B), which indicate that larger urban areas where (on average) people are more literate are picked up. This has been verified by checking the coordinate range. The dip in the Internet volume distribution (Fig 2 E) can be explained by fixed volume Internet packages that the mobile operator provides.

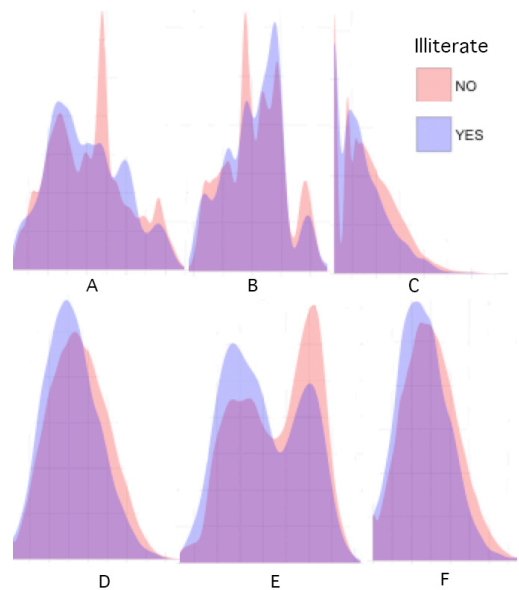

**Fig. 2.** Density distributions of top 6 mobile phone predictors of individual illiteracy A) Longitude B) Latitude C) Incoming SMS (log) D) Entropy of contacts E) Internet volume (log) F) Number of places visited

## 3.2 Geographical illiteracy mapping

A natural next step is to move from individual illiteracy to geographical illiteracy. In big Asian cities there are often thousands of mobile towers that can be used as "sensors" to estimate illiteracy rates in the areas covered by the towers. In the rural areas where towers are less dense, interpolation techniques can be utilized to include information from the neighbour towers. In figure 3 A) we have mapped out the predicted illiteracy rate per tower, in one of the larger cities.

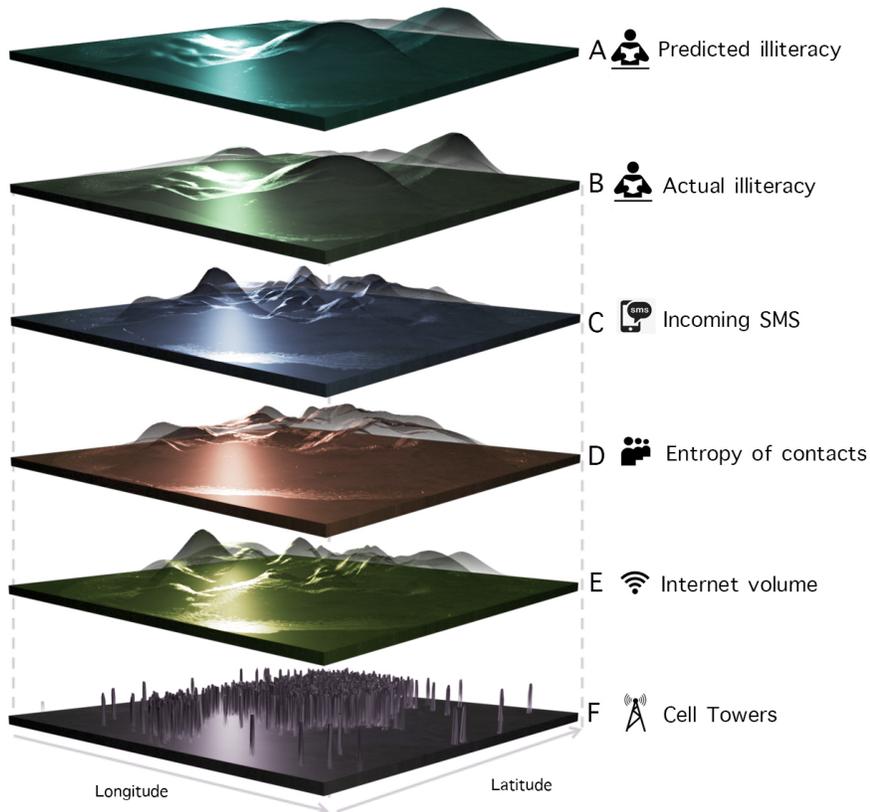

**Fig. 3.** Geographical mapping of illiteracy, top predictors and the cell tower distribution in one major citiy. Height (z-axis) is proportional to the tower averages for each given metric.

The individual illiteracy rates are here calculated by using the test set, aggregated and averaged to tower level, and then further spatially interpolated, using an IDW algorithm [19], to average out the noise of local variations between towers. The *actual* illiterate rates in Fig 3b is calculated by using the training set as ground truth. We notice three large pockets of larger illiteracy rates in the city. By also including distributions of the top predictors (Fig 3 c-e) it is possible to visually observe spatial correlations. E.g. we can observe a large area of high SMS activity (Fig 3c,left) that can be associated with low illiteracy rates. Such a visual approach might help the interpretation of findings: In most areas we will observe anti-correlated behaviour of illiteracy and the top predictors, but given the non-linearity of the GBM model, where several terms interact, this is not always the case. E.g. it has been reported from the mobile operator that some groups of illiterate people have learned to answer SMS when special mobile offers are marketed. Such learnings can quickly go viral, and spread to nearest social and spatial neighbourhood. A traditional linear model will not easily pick up these complex non-linear interactions.

## 4 Discussion and challenges

One general concern in such studies is always the sampling selection bias. A large data set may make the sampling rate irrelevant, but it doesn't necessarily make it representative [20]. The fact that the people who use mobile phone are not necessarily a representative sample of the larger population considered. This issue is especially of high relevance when considering how mobile phone data may be used for monitoring, economic forecasting and development. Research studies are often based purely on data from one mobile operator, and depending on the type of data one can expect wealthier or poorer, more males than females and uneducated or educated individuals. Even if data from all operators in a country were available, nearing the total population, it is still not the whole population. In our study we consider data from *one* large operator, where people has to own a mobile phone to be counted. We argue, however, that there should be a good correlation between *population literacy* and *sample literacy*, since the sample is large, and most people in the country own a mobile phone. Additional sources for external validation, except the two large-scale surveys, were not easily obtainable.

Another challenge is that behavioural "signals" might change over time and place. As traditional mobile communication might be transferred to other platforms (e.g. WhatsApp, Facebook messenger) the mobile phone operators might not be able to capture the social network as well as before. The model input features will therefore change accordingly. The choice of data sources and the frequency of model re-training will be vital when considering the research questions. Also, especially in development countries, people might switch between various SIM-cards to save money, and several persons might use the same phone. In countries where this is a problem, the prediction algorithm might get confused, and model performance lead to higher number of false positives. By dealing with such issues a priori, e.g. including questions around it in validation survey, we should expect an even higher accuracy.

The last challenge addresses correlation and causation. Most often the datasets are (by nature) observational and can therefore not measure causality [21]. In our study we don't explain behaviour (deductive science), rather just predicting behaviour. Illiteracy might have many causes, such as living conditions, including poverty, parents with little schooling, lack of books at home, learning disabilities, doing badly or dropping out of school. Where you live, or what cell tower you use most, do not alone decide your educational status. If an illiterate move from one area to another area, with few illiterates, this doesn't cause the person more literate. Oppositely, moving from a well-educated area into a poor neighbourhood, with overrepresented illiteracy, doesn't make a person more illiterate. As the model consists of a non-linear combination of many predictors and weightings, there is a reduced risk of such errors, and less chance of false positives or negatives: it is not one or two predictors alone that decide whether a person is illiterate. The same goes for the geographical level. If illiteracy in one area increases, we might see a drop in the incoming SMS, social entropy and Internet volumes – and the complex relationships between the 19 features in the model make the final call whether an early warning signal should be sent out. Of course there is a trade-off on accuracy between such a "black box" approach and an interpretable linear model. Consequently we do not claim to have found causal patterns between mobile phone usage and illiteracy – but potentially useful predictive signals for NGO's who need complementary estimates to detect illiteracy on a higher spatial and temporal resolution.

## 5  Conclusion

This study shows how illiteracy can be predicted from mobile phone logs, purely by investigating users' metadata. By deriving economic, social and mobility features for each mobile user we predict individual illiteracy status with 70% accuracy. Further we show how individual illiteracy can be aggregated and mapped geographically with high spatial resolution on cell tower level. Feature investigation indicates that home cell tower and incoming SMS are the superior predictors, followed by diversity of communication partners and Internet volume.

An important policy application of this work is the prediction of regional and individual illiteracy rates in underdeveloped countries where official statistics is limited or non-existing. For future work we would like to address the issues mentioned in chapter 4, and see how our method reflect the overall population illiteracy, as well as verifying the robustness of predictors in other countries. We would also like to combine mobile phone data with other spatial sources to address how poverty mapping can be done more efficiently to support the UN sustainable development goal of eradication of extreme poverty within 2030.